\documentclass[runningheads]{llncs}

\usepackage{eccv}

\usepackage{eccvabbrv}

\usepackage{graphicx}
\usepackage{booktabs}

\usepackage[accsupp]{axessibility}  

\usepackage{multirow, todonotes, siunitx, wrapfig, tikz, dsfont}
\usetikzlibrary{calc, positioning, backgrounds, arrows.meta, shapes.geometric, fit}

\usepackage{hyperref}

\usepackage{orcidlink}

\DeclareMathOperator*{\EX}{\mathop{\mathbb{E}}}

\begin{document}

\title{FLEET: Token-Based Feature Extraction for Event Camera-based Reinforcement Learning} 


\titlerunning{FLEET: Token-Based Feature Extraction for Event Camera-based RL}

\author{Tristan Gottwald\orcidlink{0009-0004-3550-9779} \and
Maximilian Schier\orcidlink{0009-0007-4314-728X} \and
Melanie Schaller\orcidlink{0000-0002-5708-4394} \and
Bodo Rosenhahn\orcidlink{0000-0003-3861-1424}}
\authorrunning{T.~Gottwald et al.}

\institute{
Institute of Information Processing, L3S, Leibniz University Hannover, Germany\\
\email{\{gottwald,schier,schaller,rosenhahn\}@tnt.uni-hannover.de}}

\maketitle

\begin{abstract}
    Event cameras generate asynchronous, high-frequency data streams offering spatially sparse information at lower latency than traditional cameras.
    In principle, these properties should be ideal for the design of control policies.
    However, reinforcement learning research in this field remains limited as existing approaches fail to fully exploit the sensor's properties.
    CNN-based methods negate the sensors benefits by aggregating events into sparse grids. This couples compute cost to sensor resolution and blurs the temporal information. Meanwhile, existing generative baselines rely on the availability of trajectory data to pretrain the model.
    We propose FLEET (Feature Learning from Events via Efficient Tokenization), a feature extractor that processes event sequences directly. Leveraging random Fourier features and cross-attention, our architecture compresses variable streams into fixed-size latent representations. 
    This decouples inference cost of the feature extractor's backbone from the sensor's resolution, enabling end-to-end learning without auxiliary losses.
    We validate FLEET on a new, high-throughput benchmark. The results demonstrate that our sequence-based approach surpasses SOTA performance and exhibits superior robustness to variations in observation frequencies.
    Code: \url{https://github.com/tgottwald/FLEET}
  
\end{abstract}

\section{Introduction}
\label{sec:intro}

In recent years, event cameras have emerged as a transformative alternative to standard frame-based sensors~\cite{shariff2024event, chakravarthi2024recent, rebecq2018emvs, gallego2017accurate}. Unlike traditional cameras that capture snapshots of the world at fixed intervals, event cameras respond asynchronously to brightness changes at the pixel level~\cite{gehrig2024low}. This paradigm offers microsecond-latency and high dynamic range, making them theoretically ideal for time-critical applications~\cite{forrai2023event, ziegler2025event} while significantly reducing the data bandwidth required for transmission~\cite{gehrig2024low}. These properties are particularly advantageous for robotics~\cite{mahlknecht2022exploring, mueggler2018continuous, forrai2023event, funk2024evetac, klenk2024deep} and autonomous driving~\cite{bulzomi2025real, cao2024embracing, shariff2024event, gehrig2021dsec, maqueda2018event}, domains characterized by highly dynamic settings where hardware is constrained and low latency is essential for safety~\cite{gehrig2024low}.
\begin{figure}[tb]
  \centering
  \includegraphics[width=0.8\textwidth]{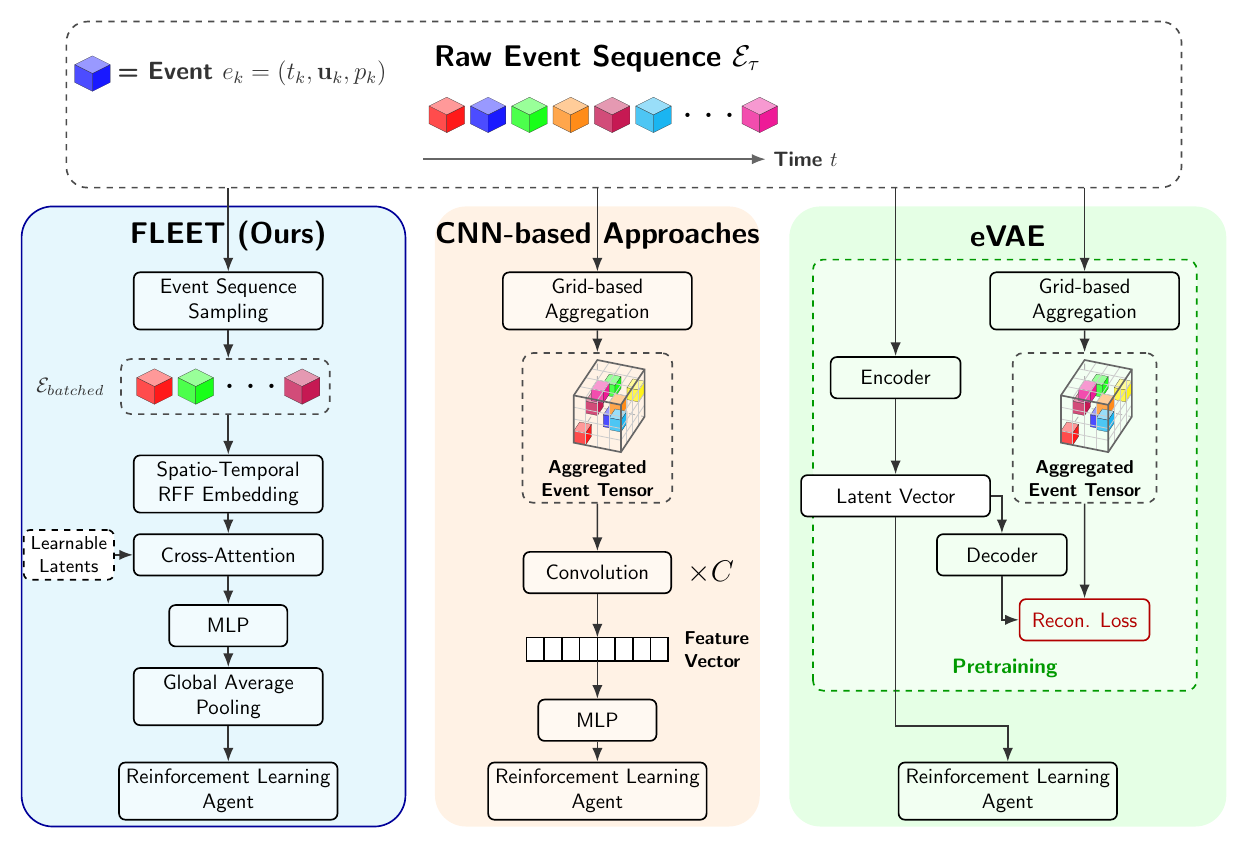}
  \caption{
  Comparison of feature extraction architectures for event camera-based reinforcement learning. Our proposed FLEET architecture (left) processes raw sequence data directly, successfully preserving temporal and spatial resolution. Events are encoded using random Fourier features and mapped to a fixed set of learnable latent vectors using cross-attention. In contrast, standard CNN-based approaches (center) force events into extremely sparse aggregated tensor representations, leading to quantization errors and unnecessary processing overhead due to sparsity. Generative pretraining pipelines like eVAE (right) attempt to solve this but suffer from an objective mismatch: pretraining on an auxiliary reconstruction task forces the representation to encode visual fidelity rather than the features actually relevant to optimal control. Furthermore, these frozen, two-stage approaches can suffer significantly from distribution shifts if the offline pretraining data fails to capture the full scope of dynamics encountered during active deployment.
  }
  \label{fig:init_img}
\end{figure}
However, realizing this potential in reinforcement learning (RL) remains a significant challenge due to the sparse, asynchronous, and irregular structure of event data. Standard deep learning pipelines typically rely on convolutional neural networks (CNN) designed for dense, synchronous images~\cite{mnih2015human, espeholt2018impala, yarats2021mastering}. To bridge this modality gap, existing approaches aggregate event streams into sparse spatio-temporal tensors~\cite{xu2024dmr, gehrig2023recurrent,walters2023ceril, zubic2024state}. We argue that this is counterproductive for control: at standard control frequencies, aggregation introduces motion blur that destroys temporal precision, while at high frequencies, it generates computationally wasteful, sparse tensors where the compute cost scales with sensor resolution rather than information content.\\
Alternative approaches employing representation learning, such as approaches based on variational autoencoders~\cite{vemprala2021representation}, attempt to compress this data into latent vectors. However, these methods rely on auxiliary reconstruction losses that prioritize visual fidelity over control-relevant features~\cite{zhang2020learning}. This creates a misalignment of objectives, where the encoder wastes capacity modeling irrelevant background dynamics rather than the task-critical physics required by the agent. Furthermore, these two-stage pipelines can suffer from distribution shift if the data used during the pretraining is not representative of the data encountered during training. In conclusion, there is a distinct lack of investigation into lightweight, end-to-end trainable architectures that can process event streams directly within the RL loop.\\
To address these limitations, we propose FLEET (Feature Learning from Events via Efficient Tokenization), a tokenized, sequence-based processing pipeline as shown in \cref{fig:init_img}. 
While the initial dimensionality reduction of the cross-attention module scales linearly with the number of input events, it maps the embedded events to a fixed-size set of latent tokens~\cite{jaegle2021perceiver, sabater2022event}. This architecture ensures that the computational cost of all subsequent processing remains independent of the input resolution or event density.
We leverage random Fourier features~\cite{rahimi2007random} to overcome the spectral bias of standard MLPs~\cite{tancik2020fourier}, enabling the network to capture high-frequency spatial details without explicit grid discretization.\\
The \textbf{main contributions} of this paper are as follows: \begin{itemize} 
    \item \textbf{Spatial Resolution-Decoupled Feature Extractor}: We present a novel, end-to-end trainable architecture that processes raw event sequences using a Perceiver-style cross-attention mechanism. By querying the variable-length event stream with a fixed set of latent queries, we decouple the computational cost of the heavy processing block from the sensor resolution, enabling efficient scaling to sensors with higher spatial resolution.
    \item \textbf{Task-Aligned Representation Learning}: Unlike generative baselines, our architecture is trained purely on the RL reward signal without auxiliary reconstruction losses. We demonstrate that this yields task-aligned representations that generalize well to distribution shifts during deployment, as the model learns to attend solely to control-relevant dynamics.
    \item \textbf{Spectral Bias Mitigation}: We integrate random Fourier features to embed coordinate inputs. We show that this allows the network to resolve high-frequency spatial details essential for obstacle avoidance, which are often lost in grid aggregations.
    \item \textbf{Frequency Robustness \& Evaluation}: We thoroughly evaluate our approach against SOTA baselines on a new Event-CarEnv benchmark based on CarEnv\cite{schier2023learned}, which enables high-throughput Event-RL research. Our results verify that our sequence-based approach outperforms SOTA approaches and exhibits lower susceptibility to variations in observation frequency compared to frame-based methods, maintaining robust performance. 
    \item The code is available at \url{https://github.com/tgottwald/FLEET}.
\end{itemize}

\section{Related Work}
Previous research combining event camera data with machine learning has predominantly focused on supervised learning tasks~\cite{sun2022ess, cao2024embracing, gehrig2023recurrent, zubic2024state, sabater2022event, zheng2023deep, hidalgo2022event}. Less attention has been given to utilizing event cameras directly as input for reinforcement learning agents~\cite{walters2023ceril, vemprala2021representation, gelada2019deepmdp}.\\
Regarding dimensionality reduction, the supervised approach most closely related to our proposed RL-based feature extractor is that of Sabater \etal~\cite{sabater2022event}. They employ a transformer-based architecture for traditional classification tasks, leveraging Perceiver-style cross-attention~\cite{jaegle2021perceiver} to reduce sequence length. Their approach uses aggregated event tensors as inputs, patchifies them and adds positional encoding~\cite{vaswani2017attention} to encode the positional information of the individual patches prior to tokenization. FLEET skips the aggregation and patchification completely and directly works on the information provided by the events. \\
In the domain of robotic control, Forrai \etal~\cite{forrai2023event} tackle an object-catching task using a quadrupedal robot. However, their RL agent does not process event data directly. Rather, classical machine learning methods estimate the object's impact point. Only these estimated coordinates are passed to the agent, restricting its role strictly to controlling the positioning of the robot.\\
Vemprala \etal~\cite{vemprala2021representation} utilize a pretrained latent space representation for obstacle avoidance using an autonomous UAV equipped with a simulated event camera. They pretrain a variational autoencoder (VAE) on manually collected data to learn an expressive latent representation of the events. Their encoder processes raw events within a specific time window via a shared multilayer perceptron embedding layer, optionally combining them with a temporal embedding. The decoder then reconstructs an event tensor, minimizing the distance to a tensor aggregated from the raw events. In contrast, our approach eliminates this pretraining phase and the associated need for offline data collection by training end-to-end directly on the control task.\\
Xu \etal~\cite{xu2024dmr} apply RL to multimodal inputs, combining RGB images with event camera data. Their model employs a contrastive loss to map fused task-relevant representations and modality-specific noise into a shared latent space, maximizing the distance between these embeddings to explicitly decouple signal from sensor-specific noise. Furthermore, they utilize the off-policy DeepMDP framework~\cite{gelada2019deepmdp} for the auxiliary task of next-state and reward prediction. Our method, conversely, operates without the need for auxiliary losses.\\
Other work, such as Walters \etal~\cite{walters2023ceril}, evaluates RL algorithms on standard, simplified environments adapted to output simulated event camera data.\\
Additionally, due to the asynchronous nature of event sensor outputs, several studies have explored Spiking Neural Networks (SNNs) for event-based RL agents \cite{zanatta2023directly, kou2025event, xu2024dmr}. However, SNNs remain notoriously difficult to train and often require specialized neuromorphic hardware~\cite{deng2020rethinking, pfeiffer2018deep, bulzomi2025real}, limiting their general applicability.

\section{Fundamentals}
In this section we present the foundations of event cameras and Markov decision processes.
\subsection{Event Cameras}
Standard cameras capture images synchronously at a constant sampling frequency (\eg \SI{30}{\hertz}), recording all pixels simultaneously~\cite{gallego2020event}. In contrast, event cameras~\cite{lichtsteiner2008128} operate asynchronously.  They measure per-pixel brightness changes and output an event only when the luminance change exceeds a specific threshold~\cite{gallego2020event, chakravarthi2024recent}.
This generates an asynchronous, high temporal resolution stream of events, denoted as $\mathcal{E}=\{e_k \mid k=1,\ldots, \mathcal{K}\}$. Each  event $e_k$ is defined by a tuple $(t_k,\mathbf{u}_k,p_k)$, where $t_k$ is the timestamp of the change, $\mathbf{u}_k =(x_k, y_k)$ specifies the triggered pixels coordinates and $p_k$ indicates the polarity of the event (positive or negative change in luminance)~\cite{zheng2023deep}.\\
Transmitting this asynchronous stream instead of images addresses the central bandwidth-latency tradeoff~\cite{gehrig2024low} found in time-critical applications. 
For standard RGB cameras, one must compromise: achieving low latency and high temporal resolution requires high frame rates that saturate bandwidth with highly redundant data, whereas lowering the frame rate to conserve bandwidth increases latency and risks missing critical inter-frame dynamics.
However, this asynchronicity introduces significant processing challenges~\cite{gallego2020event}, as classical computer vision algorithms are designed for synchronous data formatted in dense grid structures~\cite{han2022survey, deng2009imagenet, zhu2024vision}. To bridge this gap, various event stream representations have been developed to make the data compatible with existing vision-based architectures~\cite{gehrig2023recurrent, zheng2023deep, maqueda2018event, zihao2018unsupervised}.\\
To process asynchronous events with standard vision architectures, existing methods frequently aggregate the continuous event stream into dense grid structures~\cite{gehrig2023recurrent,xu2024dmr, zubic2024state }. A standard approach~\cite{gehrig2023recurrent} discretizes events over a sampling window $\tau$ into a spatiotemporal tensor of size $H \times W \times T \times 2$. In this format, $H$ and $W$ denote spatial resolution, $T$ represents discrete temporal bins, and the final dimension captures event polarity. By flattening the temporal and polarity dimensions, the data becomes compatible with classical CNNs. However, this forced discretization  yields highly sparse tensors that are computationally inefficient to process~\cite{perot2020learning} and tightly couples the computational cost to the sensor's resolution rather than the actual information content.

\subsection{Markov Decision Process}
Reinforcement learning is formalized as a Markov decision process (MDP), which is defined as a five-element tuple $(\mathcal{S}, \mathcal{A}, P, r, \gamma)$~\cite{sutton1998reinforcement, bellman1957markovian}. $\mathcal{S}$ represents the state space, $\mathcal{A}$ is the action space. The dynamics model $P: \mathcal{S} \times \mathcal{A} \rightarrow \mathcal{P}(\mathcal{S})$ describes the transition probability $P(\mathbf{s}_{t+1}|\mathbf{s}_t, \mathbf{a}_t)$ of the agent reaching state $\mathbf{s}_{t+1} \in \mathcal{S}$ after executing action $\mathbf{a}_t \in \mathcal{A}$ in its current state $\mathbf{s}_{t} \in \mathcal{S}$.
Typically this definition is extended with the initial state distribution $\rho_0: \mathcal{P}(\mathcal{S})$, which indicates in which states the agent may start its trajectory. A trajectory $\zeta = (\mathbf{s}_0, \mathbf{a}_0, \mathbf{s}_1, \mathbf{a}_1, \ldots, \mathbf{s}_\mathcal{T}, \mathbf{a}_\mathcal{T},)$ describes a consecutive sequence of states and actions. For each transition the agent is awarded with a reward according to the reward function $r: \mathcal{S} \times \mathcal{A} \times \mathcal{S}\rightarrow \mathbb{R}$.
Based on the rewards received and the discount factor $\gamma$, the discounted future return is defined as $G(\zeta) = \sum_t \gamma^tr_t$.
The agent's goal is to find an optimal policy $\pi^*$ which maximizes the expected return $J_r$:
\begin{equation} \label{eq:mdp_optimal_policy}
    \pi^* = \arg\max_{\pi}J_r(\pi)\text{, with } J_r(\pi)=\EX_{\zeta\sim\pi}[G(\zeta)] \, .
\end{equation}
A policy $\pi: \mathcal{S} \rightarrow \mathcal{P}(\mathcal{A})$ expresses the probability of selecting action $\mathbf{a} \in \mathcal{A}$ while the agent is in state $\mathbf{s} \in \mathcal{S}$. The shorthand notation $\zeta\sim\pi$ indicates that the trajectory $\zeta$ is obtained by a random rollout of $\pi$.

\section{Proposed Method}
\begin{figure}[tb]
  \centering
  \includegraphics[width=1.0\textwidth]{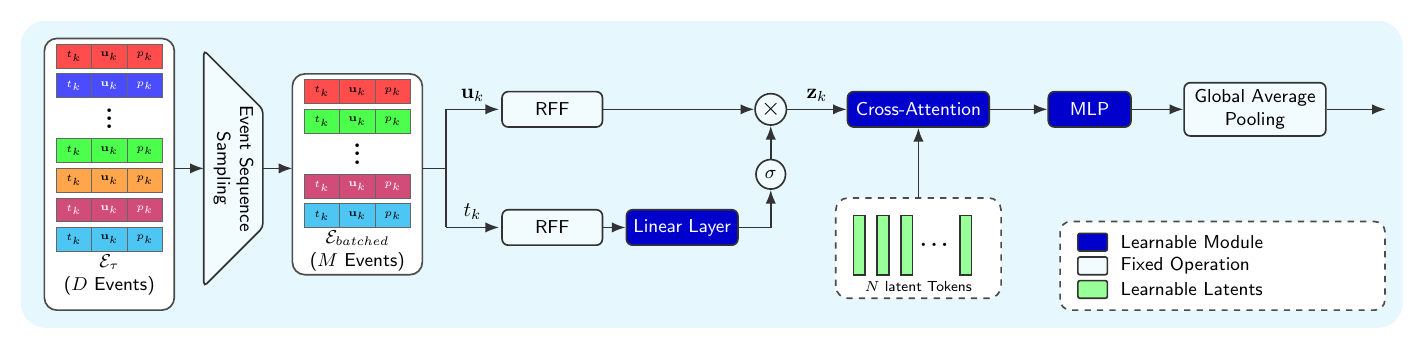}
  \caption{Overview of our proposed FLEET feature extractor. The number of events within the original sequence $\mathcal{E}_\tau$ is first reduced or padded to $\mathcal{E}_\text{batched}$. The temporal and positional component of each event in $\mathcal{E}_\text{batched}$ are embedded using shared temporal and spatial random Fourier feature encoders. These embeddings are combined to $\mathbf{z}_k$ using a temporal gating mechanism. The cross-attention module maps the sequence of embedded events to $N$ learnable latent vectors to reduce the dimensionality. The cross-attention is followed by an MLP encoder block. As a last step global average pooling is performed over all latents to form a final representation. 
  }
  \label{fig:event_seq_enc}
\end{figure}
We propose Feature Learning from Events via Efficient Tokenization (FLEET), a simple yet effective feature extractor architecture. FLEET learns an expressive latent representation directly from raw event camera data for reinforcement learning.
Our feature extractor avoids aggregating events into sparse grid structures and requires neither pretraining nor auxiliary tasks.\\
Instead of aggregating all temporally sorted events within the event stream $\mathcal{E}_\tau$ over a sampling period $\tau$ into a sparse tensor, FLEET operates directly on a subsampled, fixed-length sequence, denoted as $\mathcal{E}_\text{batched}$.
To reduce the original sequence $\mathcal{E}_\tau$ of length $D$ to a fixed length $M$, $\mathcal{E}_\tau$ is divided into $M$ equally sized bins. The bin edges are defined by the indices $b_k = \lfloor k\frac{D}{M} \rfloor, k \in {0, \dots, M-1}$, with $b_M = D$. For each bin, a single event at a random index $c_k \sim \mathcal{U}\{b_{k}, b_{k+1}-1\}$ is selected.
If there are fewer than $M$ events, $\mathcal{E}_\text{batched}$ is padded with zero values which are masked in subsequent operations. 
By operating on this fixed-length sequence, our architecture remains closely aligned with the asynchronous paradigm of event cameras. This bypasses the need for the computationally expensive CNNs~\cite{gehrig2023recurrent, zubic2024state, xu2024dmr} typically required to overcome the sparsity of grid-based representations.

\subsubsection{Event Tokenization and Embedding}
The temporal $t \in \mathbb{R}$ and spatial components $\mathbf{u} \in \mathbb{R}^2$ of the events are normalized to $[-1,1]$ and projected into a higher-dimensional embedding space $d_{\text{embed}}$. Standard MLPs suffer from a spectral bias that smooths over high-frequency details from low-dimensional inputs~\cite{tancik2020fourier}. To overcome this bias and preserve fine-grained spatio-temporal structures, such as small obstacles or rapid scene dynamics, the coordinates are mapped using random Fourier features (RFF)~\cite{rahimi2007random}:
\begin{equation} \label{eq:_rff}
    \text{RFF}(\mathbf{x})=[\sin(\pi\mathbf{x}B),\cos(\pi\mathbf{x}B)],
\end{equation}
where $B \in \mathbb{R}^{d_\text{input} \times d_{\text{embed}}/2}$ is sampled from a Gaussian distribution
\begin{equation} \label{eq:_rff_b}
    B_{ij} \sim \mathcal{N}(0, \sigma^2)\,.
\end{equation}
The parameter $\sigma$ scales the frequency spectrum of the features.
Tancik \etal have shown that this scale parameter is essential for learning high frequency data like coordinates~\cite{tancik2020fourier}, thus we choose different scalings $\sigma_{\text{temporal}}$, $\sigma_{\text{spatial}}$ for the temporal and spatial components with the former being chosen to be smaller than the latter.\\
To effectively fuse the spatiotemporal dynamics, we introduce a temporal gating mechanism. The temporal embedding is linearly projected via learned parameters $W_\text{gate}$ and $\mathbf{b}_\text{gate}$, and passed through a sigmoid activation. This generates a temporal modulation vector that gates the spatial embedding via an element-wise Hadamard product:
\begin{equation} \label{eq:embed}
    \mathbf{z}_{k} =  \text{RFF}(\mathbf{u}_k) \otimes \text{sigmoid}(W_\text{gate}\text{RFF}(t_k) + \mathbf{b}_\text{gate})\,.
\end{equation}
\subsubsection{Event Sequence Condensation}
To decouple downstream computational load from the sequence length $M$, we employ a Perceiver-style~\cite{jaegle2021perceiver} condensation module using cross-attention~\cite{vaswani2017attention}:
\begin{equation} \label{eq:cross_attn}
    \text{cross-attention}(Q,K,V)=\text{softmax}\left(\frac{QK^T}{\sqrt{d_k}}\right)V\,.
\end{equation}
For dimensionality reduction, $N$ learnable latent vectors form the query matrix $Q\in \mathbb{R}^{N\times d_{\text{embed}}}$, while the embedded event sequence $\mathcal{Z}$ provides the keys $K$ and values $V \in \mathbb{R}^{M\times d_{\text{embed}}}$. In practice, we use multi-head cross-attention across $h$ heads, applying a padding mask $M_{\text{mask}} \in \mathbb{R}^{N \times M}$ before the softmax to ignore padded values. This bottleneck compresses the variable-density stream into a dense, fixed-size representation by dynamically attending to the most relevant events.

\subsubsection{Backbone and RL Integration}
The $N$ condensed tokens are processed by a three-layer MLP encoder to capture global scene context and aggregated via global average pooling into a single continuous feature vector. Our FLEET feature extractor passes this representation to a standard Proximal Policy Optimization (PPO) agent \cite{schulman2017proximal}. Crucially, while the actor and critic share the feature extractor, we apply a stop-gradient operation to the actor's inputs \cite{kostrikov2020image}. Consequently, the feature extractor's parameters are updated exclusively via the critic's value function loss. Driving the representation learning purely through temporal difference updates avoids objective conflicts between the actor and critic, stabilizing the training process.

\section{Experiments}
In this section, FLEET is evaluated and compared to multiple baseline feature extractors. Our experiments follow best practice~\cite{agarwal2021deep} and report the interquartile mean (IQM) and 95\% confidence interval (CI) using 50\,000 bootstrap iterations across ten randomly seeded runs. The parameters for the PPO agent and the individual feature extractors can be found in the Appendix.

\subsection{Event-CarEnv} \label{sec:event_carenv}
\begin{figure}[tb]
  \centering
  \begin{subfigure}{0.45\linewidth}
    \centering
    \includegraphics[width=1.0\textwidth]{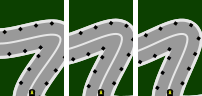}
    \caption{RGB Camera View}
    \label{fig:car_env_rgb}
  \end{subfigure}
  \hfill
  \begin{subfigure}{0.45\linewidth}
    \centering
    \includegraphics[width=1.0\textwidth]{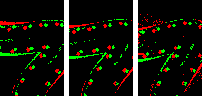}
    \caption{Event Camera View (accumulated)}
    \label{fig:car_env_event}
  \end{subfigure}
  \caption{Sequence of RGB (\cref{fig:car_env_rgb}) and event (\cref{fig:car_env_event}) camera observations in the Event-CarEnv Racing scenario. Note that the event camera data has been temporarily aggregated into an image format strictly for human interpretability.}
  \label{fig:car_env}
\end{figure}

Prior research combining event cameras with reinforcement learning~\cite{vemprala2021representation, xu2024dmr} has predominantly relied on the Microsoft AirSim~\cite{shah2017airsim} and CARLA~\cite{Dosovitskiy17} simulators. While these platforms excel at photorealistic simulation, their underlying architectures prioritize visual fidelity over the rapid, high-throughput environment interactions required for algorithmic iteration in on-policy RL.
To isolate the core mechanics of event-based policy learning, we selected the CarEnv~\cite{schier2023learned} simulator as the foundation for our experiments. CarEnv is explicitly designed to facilitate RL research, providing a highly dynamic, control-focused setting that mirrors the control complexity of established real-world autonomous driving benchmarks, such as Duckietown~\cite{paull2017duckietown} and the IEEE Bosch Future Mobility Challenge~\cite{kilyen2021ieee}.
Our experiments focus on CarEnvs Racing and Parking scenario.
In the Racing scenario, the agent controls a vehicle on a procedurally generated circuit and is rewarded for driving quickly while avoiding track boundaries. This domain randomization frequently introduces sharp turns, forcing the agent to learn a robust policy capable of generalizing across varied track layouts.\\
The Parking scenario consists of a randomized parking bay where the agent must safely align the vehicle with a target parking position without colliding with the boundaries. Previously, observations within CarEnv have been restricted to set-~\cite{SchSch2025a, farkas2025residual, cramer2024contextualized} or lidar-based~\cite{gottwald2024safe} modalities and their respective feature extractors.\\
We extend CarEnv with standard RGB and event camera sensors. Racing uses a \SI{30}{m} frontal lookahead, while Parking centers the vehicle with a \SI{5}{m} front-and-rear range to enable tight backward and lateral maneuvers. Both sensors use an identical $96 \times 64$ pixel spatial resolution, ensuring a fair comparison and matching the low spatial resolution of standard visual RL benchmarks~\cite{mnih2015human,espeholt2018impala}. We model event camera dynamics similarly to \cite{rebecq2018esim}, estimating asynchronous continuous-time events from discrete RGB renderings via stochastic logarithmic luminance thresholds. See the Appendix for the full formulation.\\
The event observation $\mathbf{x}_{\text{event}}$ is augmented with a state vector $\mathbf{x}_{\text{state}}$. This comprises velocities $v_x$ and $v_y$, yaw velocity $\omega$, and steering angle $\beta$ for Racing, and is limited to $\mathbf{x}_{\text{state}}=(v_x, \beta)$ for Parking~\cite{schier2023learned}.
\cref{fig:car_env} illustrates a sequence of RGB images alongside aggregated event camera data for the Racing scenario.

\subsection{Event Inverted Double Pendulum}
To further evaluate FLEET on a classic continuous control benchmark, we adapt the Inverted Double Pendulum task~\cite{barto2012neuronlike, towers2024gymnasium}. We integrate the same event generation pipeline as used in Event-CarEnv, configuring the simulated event camera to a spatial resolution of $128 \times 128$ pixels. 
Complete details regarding $\Delta t=0.01$, episode lengths and reward scaling are provided in the Appendix.

\subsection{Baselines}
Given the scarcity of event-based RL, we compare FLEET against the two most recent reproducible baselines~\cite{vemprala2021representation, xu2024dmr} alongside additional baselines inspired by image- and set-based RL:

\begin{itemize}
\item \textbf{Dense RGB CNN:} A NatureCNN \cite{mnih2015human} trained on noise-free RGB observations with identical resolution and field-of-view. This is included to relate to the performance of frame-based RL on the problems.

\item \textbf{Grid-based CNNs:} These methods process events aggregated into RVT \cite{gehrig2023recurrent} with five temporal bins:
\begin{itemize}
    \item \textbf{NatureCNN~\cite{mnih2015human}:} Adapted from Mnih \etal with the first-layer stride reduced from 4 to 1. This prevents the network from skipping sparse, single-pixel events common in tensor-aggregated event data.
    \item \textbf{ImpalaCNN~\cite{espeholt2018impala}:} A deeper architecture using residual connections and max pooling, which typically outperforms NatureCNN in pixel-based RL.
    \item \textbf{DMRCNN~\cite{xu2024dmr}:} The event data encoder from Xu \etal, utilizing specialized kernels and terminal layer normalization for stabilization.
\end{itemize}

\item \textbf{Sequence-based Methods:} These process event streams directly:
\begin{itemize}
    \item \textbf{LFF-DS~\cite{schier2023learned}:} Based on Schier \etal but modified with FLEET's normalization and temporal gating to isolate the impact of the cross-attention.
    \item \textbf{eVAE~\cite{vemprala2021representation}:} A VAE pretrained on a reconstruction task using collected trajectory data. Since we assume that there are no demonstrations or existing policies available we use random policy rollouts.
    eVAE internally concatenates a stack of the three most recent latent representations.
\end{itemize}
\end{itemize}

\subsection{Feature Extractor Comparison}\label{sec:fe_comparison}
\begin{figure}[tb]
  \centering
  \begin{subfigure}{0.45\linewidth}
    \centering
    \includegraphics[width=1.0\textwidth]{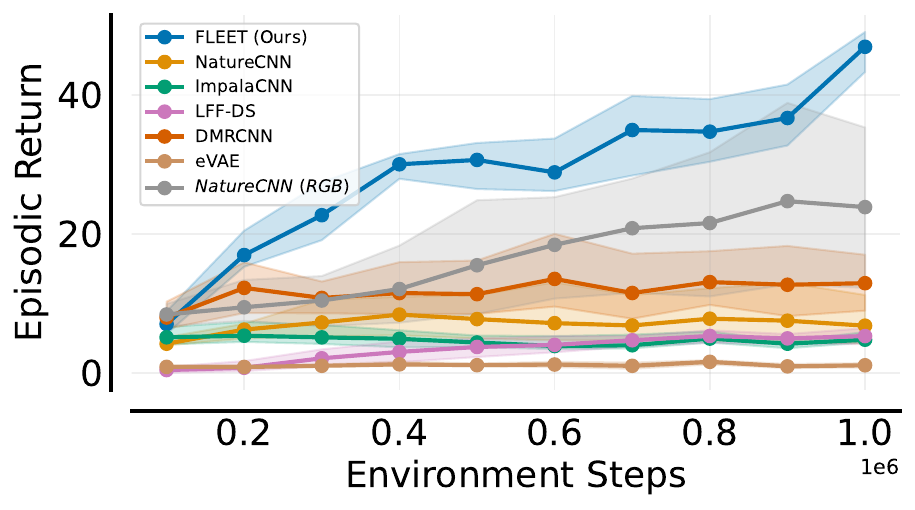}
    \caption{Event-CarEnv Racing}
    \label{fig:results_racing}
  \end{subfigure}
  \hfill
  \begin{subfigure}{0.45\linewidth}
    \centering
    \includegraphics[width=1.0\textwidth]{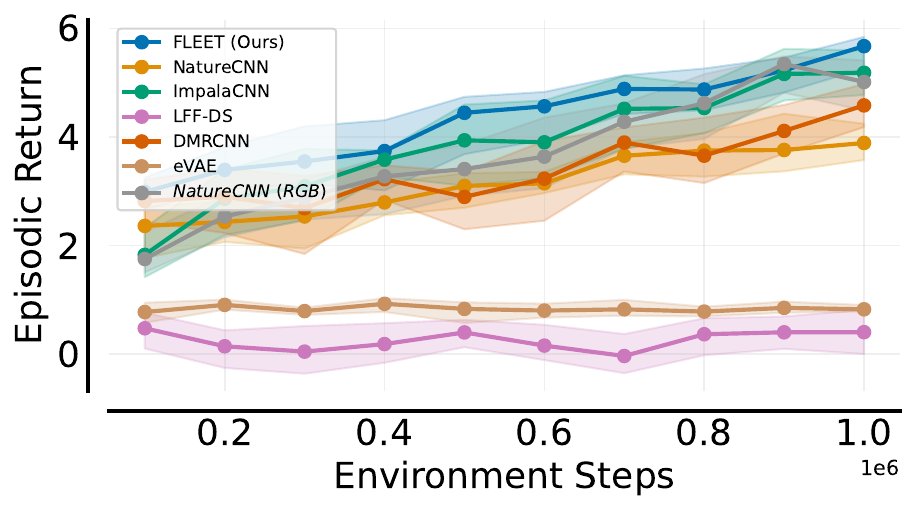}
    \caption{Event-CarEnv Parking}
    \label{fig:results_parking}
  \end{subfigure}
  \begin{subfigure}{0.45\linewidth}
    \centering
    \includegraphics[width=1.0\textwidth]{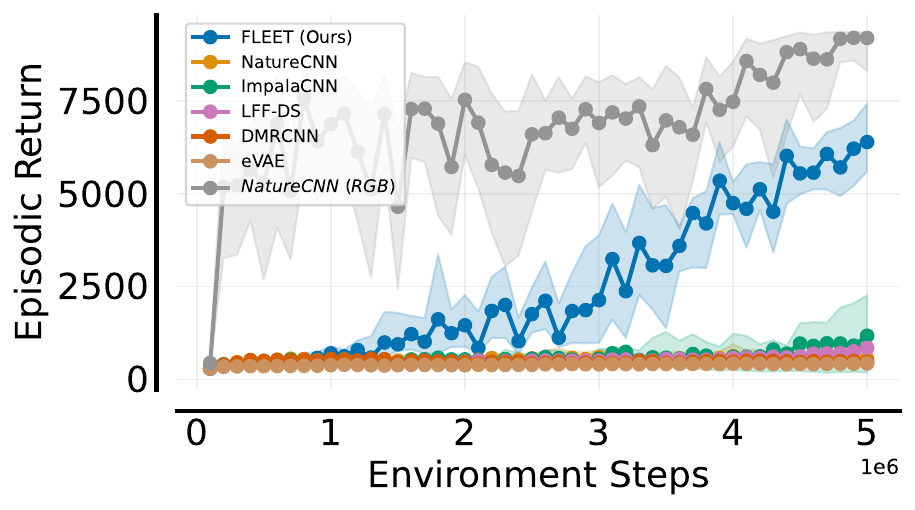}
    \caption{Event Inverted Double Pendulum}
    \label{fig:results_idp}
  \end{subfigure}
  \caption{
  Progression of the IQM and 95\% CI of the episodic return in the Event-CarEnv Racing and Parking ($\Delta t=0.05$, $10^6$ steps) and the Inverted Double Pendulum ($\Delta t=0.01$, $5\cdot10^6$ steps) settings.
  Agents were evaluated for 10 episodes every $10^5$ environment steps, reporting the mean episodic return. In the Racing scenario, CNN-based competitors fail to demonstrate learning progress, though they manage to learn suitable policies in the denser Parking setting. The agents using the eVAE and LFF-DS feature extractor fail to learn a viable policy in any environment. Across all tasks, FLEET shows superior performance to all event-based baseline methods.
  }
  \label{fig:eval_racing}
\end{figure}

We evaluate FLEET in the Event-CarEnv Racing and Parking scenarios ($\Delta t=0.05$, $10^6$ steps) and the Inverted Double Pendulum (IDP) task ($\Delta t=0.01$, $5\cdot10^6$ steps). \cref{tab:performance_ppo} and \cref{fig:eval_racing} report the final interquartile means, 95\% confidence intervals, and training progressions.
\begin{table}[tb]
\caption{
Performance of various feature extractors in combination with PPO. Interquartile means (IQM) and 95\% confidence intervals (CI95) of episodic returns are reported for Event-CarEnv Racing/Parking ($\Delta t=0.05$, $10^6$ training steps) and Inverted Double Pendulum ($\Delta t=0.01$, $5\cdot10^6$ steps). \textbf{Best} and \underline{second-best} results are highlighted. FLEET outperforms all baselines across all environments, including a RGB variant on Racing and Parking.
}
\label{tab:performance_ppo}
    \centering
    \resizebox{\columnwidth}{!}{%
        \begin{tabular}{l | r @{\,\,} r @{\,} r | r @{\,\,} r @{\,} r | r @{\,\,} r @{\,} r} \toprule
            \multirow{2}{*}{Feature Extractor} & \multicolumn{3}{c|}{Racing (Return $\uparrow$)} & \multicolumn{3}{c|}{Parking (Return $\uparrow$)} & \multicolumn{3}{c}{Inv. Dbl. Pendulum (Return $\uparrow$)} \\
             & IQM & \multicolumn{2}{c|}{CI95} & IQM & \multicolumn{2}{c|}{CI95} & IQM & \multicolumn{2}{c}{CI95} \\ \midrule
            \textit{NatureCNN (RGB)}~\cite{mnih2015human} & 23.89 & [12.71, & 35.89] & 5.01 & [4.50, & 5.41] & 8980.69 & [7883.73, & 9352.29] \\ \midrule
            NatureCNN~\cite{mnih2015human} & 6.83 & [4.41, & 11.44] & 3.89 & [3.50, & 4.82] & 560.50 & [469.09, & 1013.37] \\ 
            ImpalaCNN~\cite{espeholt2018impala} & 4.81 & [4.39, & 5.93] & \underline{5.18} & [4.77, & 5.59] & \underline{1175.57} & [184.60, & 2178.99] \\ 
            LFF-DS~\cite{schier2023learned} & 5.36 & [4.22, & 6.46] & 0.40 & [0.01, & 0.81] & 848.29 & [788.83, & 919.08] \\            
            eVAE~\cite{vemprala2021representation} & 1.13 & [0.81, & 1.43] & 0.82 & [0.78, & 0.90] & 429.67 & [414.25, & 441.36] \\ 
            DMRCNN~\cite{xu2024dmr} & \underline{12.94} & [9.09, & 16.99] & 4.58 & [4.18, & 4.96] & 470.68 & [395.26, & 504.51] \\ 
            FLEET (\textbf{ours}) & \textbf{46.96} & [43.28, & 49.03] & \textbf{5.67} & [5.24, & 5.84] & \textbf{6390.41} & [5589.50, & 7410.47] \\ \bottomrule 
        \end{tabular}
    }
\end{table}
In the Racing scenario, CNN-based baselines fail to learn viable policies from the sparse tensor representation. The strongest baseline (DMRCNN) achieves only 27\% of FLEET's final performance. Qualitatively, these low return ranges correspond to a degenerate policy where the vehicle simply drives in a straight line and crashes at the first turn. FLEET, however, demonstrates almost monotonic learning and surpasses the RGB variant by 50\%.
The set-based LFF-DS is also incapable of learning a suitable policy, underscoring the importance of FLEET's cross-attention module.
Furthermore, eVAE fails entirely, likely due to a severe distribution shift, as its pretraining on random rollouts inadequately captures the dynamic state distribution needed for robust control.
In the Parking scenario, the performance gap narrows. ImpalaCNN achieves 91\% of FLEET's episodic return, with NatureCNN reaching 68\%. We attribute this relative improvement to the smaller spatial distance covered by the sensor's field of view in the Parking setting. At an identical sensor resolution, the Racing scenario covers three times the vertical distance of the Parking scenario. This reduced spatial scale in the Parking task increases the average event density of the RVT Tensor from 9\% to 15\%.  Because the data is less sparse, the CNN-based feature extractors are less penalized for their architectural limitations.
Finally, in the IDP task, FLEET substantially outperforms all event-based competitors (6390.41 vs. <1200). However, the dense RGB baseline scores higher (8980.69). Since IDP features a static camera where the full state remains visible, it inherently favors dense frame-based sensing when bandwidth is unrestricted.

\subsection{Resolution Invariance} \label{sec:res_invariance}
A core advantage of FLEET is its ability to scale to high-resolution sensors without sacrificing control performance. As demonstrated in \cref{tab:resolution_comparison} for the Event-CarEnv Racing scenario, FLEET maintains consistently high episodic returns as the spatial resolution of the simulated event camera scales from $96\times64$ up to $384\times256$. By modeling the event stream as a fixed-length point cloud compressed into $N$ latent tokens, the architecture effectively extracts fine-grained dynamics regardless of the native sensor resolution.
We contrast this with NatureCNN and DMRCNN, the highest-performing baselines from \cref{sec:fe_comparison}. DMRCNN drops marginally while NatureCNN shows improvement at intermediate resolutions. However, this comes at a computational cost. Because CNNs process grids, their memory and compute requirements scale quadratically.
FLEET bypasses both this representation degradation and the computational bottleneck entirely. A detailed analysis of performance and resource scaling is provided in the Appendix.
\begin{table}[tb]
\caption{
Interquartile means and 95\% confidence intervals of the episodic return of FLEET for different spatial sensor resolutions, evaluated on the Racing scenarios. FLEET's performance is near constant despite an increase in spatial resolution.}
\label{tab:resolution_comparison}
    \centering
    \resizebox{\columnwidth}{!}{%
        \begin{tabular}{l|r @{\,\,} r @{\,} r |r @{\,\,} r @{\,} r |r @{\,\,} r @{\,} r |r @{\,\,} r @{\,} r} \toprule
            \multirow{3}{*}{Feature Extractor} & \multicolumn{3}{c|}{$96\times 64$}  & \multicolumn{3}{c|}{$128\times 96$}  & \multicolumn{3}{c|}{$192\times 128$} & \multicolumn{3}{c}{$384\times 256$} \\
            & \multicolumn{3}{c|}{(Return $\uparrow$)} & \multicolumn{3}{c|}{(Return $\uparrow$)} & \multicolumn{3}{c|}{(Return $\uparrow$)} & \multicolumn{3}{c}{(Return $\uparrow$)}\\
             & IQM &  \multicolumn{2}{c|}{CI95} & IQM &  \multicolumn{2}{c|}{CI95} & IQM &  \multicolumn{2}{c|}{CI95} & IQM &  \multicolumn{2}{c}{CI95} \\\midrule
             NatureCNN &  6.83 & [4.41, & 11.44] & \underline{11.43} & [9.18, & 14.70]& \underline{16.42} & [11.88,& 21.25] & \underline{13.21} & [10.64,& 17.55]\\
             DMRCNN & \underline{12.94} & [9.09, & 16.99] & 9.55 & [6.90, & 13.02]& 10.46 & [8.08,& 13.88] & 8.57 & [6.41 ,& 12.62]\\
            FLEET (\textbf{ours})& \textbf{46.96} &[43.28,& 49.03] & \textbf{40.48} &[36.49,& 44.17]& \textbf{43.65} & [36.85,& 49.64] & \textbf{44.32} & [39.89,& 49.57]\\\bottomrule 
        \end{tabular}
    }
\end{table}

\subsection{Robustness to Input and Control Frequency} \label{sec:freq}
High-frequency operation is critical for low-latency control, yet decreasing the observation period $\Delta t$ reduces the information density and captures less of the environment dynamics per sample.
We evaluate robustness by varying frequencies in the Racing scenario ($\Delta t \in \{0.1, 0.05, 0.025\}$).
To ensure a fair comparison, training steps are scaled inversely with $\Delta t$ to $\{0.5\cdot10^6, 10^6, 2\cdot10^6\}$ steps to keep the environment time spent on training consistent.
Furthermore the discount factor is adjusted to maintain consistent future reward discounting.
\begin{wrapfigure}{r}{0.52\linewidth}
    \centering
    \includegraphics[width=0.5\textwidth]{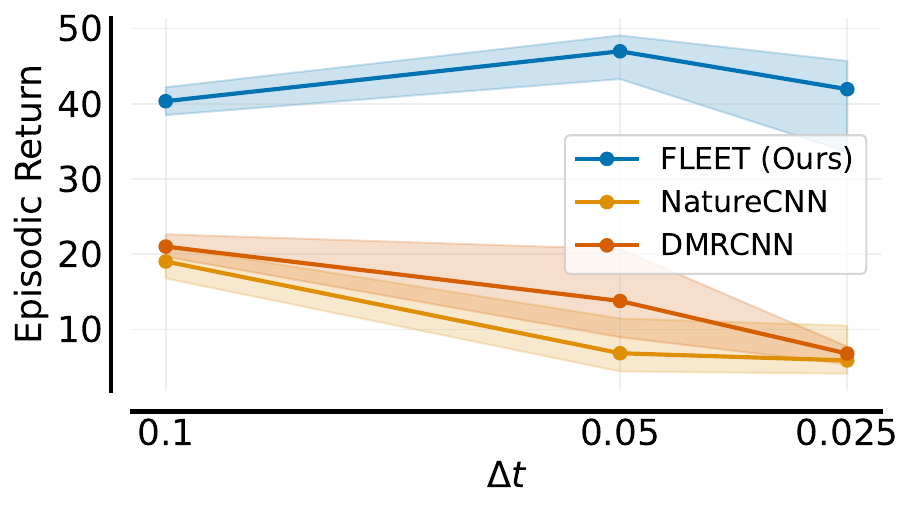}
      \caption{
        Episodic return across varying simulation step sizes ($\Delta t = \{0.1, 0.05, 0.025\}$) in the Racing scenario. FLEET remains robust at higher control frequencies, whereas CNN-based architectures degrade.  
      }
      \label{fig:eval_delta_t}
\end{wrapfigure}
As shown in \cref{tab:delta_t_comparison} and \cref{fig:eval_delta_t}, the performance of CNN-based NatureCNN and DMRCNN degrades as frequency increases.
At $\Delta t=0.025$, grid-based representations like RVT become highly sparse that standard convolutions fail to process.
In contrast, FLEET remains robust across all settings. 
By operating directly on event sequences, FLEET natively exploits fine-grained temporal dynamics rather than being bottlenecked by sparse spatial tensors. This allows our architecture to maintain high performance even as the individual observation windows shrink at higher frequencies.
\begin{table}[tb]
\caption{
Impact of simulation step size ($\Delta t \in \{0.1, 0.05, 0.025\}$) on episodic return (IQM and 95\% CI) in the Racing scenario. Training steps are scaled to $\{0.5\cdot10^6, 10^6, 2\cdot10^6\}$ respectively to preserve total environment time. FLEET maintains robust performance at higher frequencies compared to CNN-based baseline methods.
}
\label{tab:delta_t_comparison}
    \centering
    \resizebox{\columnwidth}{!}{%
        \begin{tabular}{l| r @{\,\,} r @{\,} r |  r @{\,\,} r @{\,} r |  r @{\,\,} r @{\,} r} \toprule
            \multirow{3}{*}{Feature Extractor} & \multicolumn{3}{c|}{$0.5\cdot10^6 \text{ steps @ }\Delta t=0.1$}  & \multicolumn{3}{c|}{$10^6 \text{ steps @ } \Delta t=0.05$}  & \multicolumn{3}{c}{$2\cdot10^6 \text{ steps @ }\Delta t=0.025$} \\
            & \multicolumn{3}{c|}{(Return $\uparrow$)} & \multicolumn{3}{c|}{(Return $\uparrow$)} & \multicolumn{3}{c}{(Return $\uparrow$)} \\
             & IQM & \multicolumn{2}{c|}{CI95} & IQM & \multicolumn{2}{c|}{CI95} & IQM & \multicolumn{2}{c}{CI95} \\\midrule
            NatureCNN~\cite{mnih2015human}  & 19.02 & [16.82,& 21.08] & 6.83 & [4.41,& 11.44] & 5.87 & [4.10,& 10.40] \\
            DMRCNN~\cite{xu2024dmr}  & \underline{21.00} & [19.70,& 22.62] & \underline{12.94} &[9.09,& 16.99] & \underline{6.78} & [5.36,& 7.71] \\ 
            FLEET (\textbf{ours})& \textbf{40.33} &[38.43,& 42.10] & \textbf{46.96} &[43.28, & 49.03]& \textbf{41.92} & [34.18,& 46.68]\\\bottomrule 
        \end{tabular}
    }
\end{table}

\subsection{Latent Interpretability} \label{sec:interpretability}
\begin{figure}[tb]
  \centering
  \includegraphics[width=1.0\textwidth]{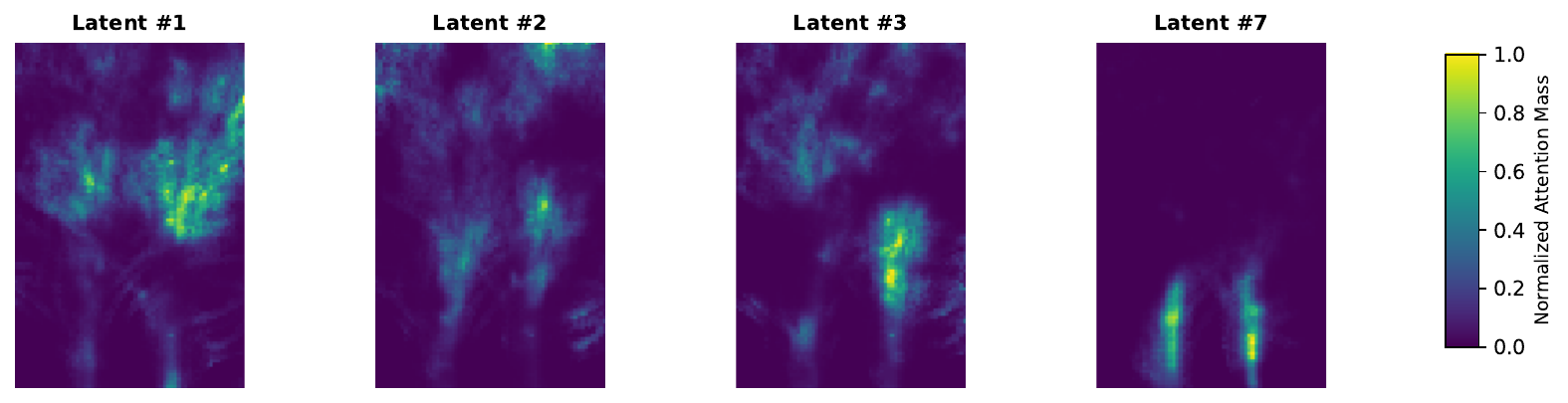}
  \caption{
  Normalized spatial attention weights of the event-to-latent mapping during a Racing episode. The activation patterns demonstrate how distinct latents specialize in different regions of the perceptive field. Latent 7 localizes on the track boundaries and cones immediately in front of the ego-vehicle, while Latent 1 attends to broader, distant dynamics in the upper frame. Because the agent maintains a centered racing line, the spatial variance of attended events naturally scales with distance. Latent 3 highlights the agent's ability to detect track curvature by jointly attending to diagonally opposed regions (\eg upper-left and lower-right), providing context for the steering policy.
  }
  \label{fig:interpretability}
\end{figure}

In this section, we investigate the interpretability of the learned latent tokens by visualizing their accumulated attention weights over an Event-CarEnv Racing episode. To ensure visual clarity, the attention mass is normalized independently for each latent. \cref{fig:interpretability} shows four representative latents, demonstrating how the cross-attention module spatially disentangles the visual field into representations for the agent. Latent 7 acts as a localized feature detector for the ego-vehicle's immediate proximity. As the agent learns to center the vehicle, this attention mass reliably forms two straight boundary lines. In contrast, Latent 1 captures distant events in the upper visual field, where changes in the camera's perspective lead to a broader, higher-variance attention distribution. Finally, Latents 2 and 3 integrate both near and distant event clusters. This hierarchical receptive field allows the agent to approximate track curvature, which is a crucial geometric feature for learning a robust steering policy.

\subsection{Ablation}
Due to page limitations we leave the ablations on different sequence sampling approaches, the number of latent tokens $N$, the maximum sequence length $M$ and embedding aggregation techniques to the Appendix and focus on the core architectural decisions of FLEET.
\paragraph{Effect of Components:}
We systematically ablate FLEET’s core modules (\cref{tab:ablation_module}). Replacing the cross-attention bottleneck with self-attention increases complexity from $\mathcal{O}(NM)$ to $\mathcal{O}(M^2)$ (where $M \gg N$) and severely degrades performance. Substituting the MLP backbone with a Transformer encoder drops performance by 24\%. Unlike prior methods \cite{vemprala2021representation}, FLEET strongly leverages temporal information. Removing temporal embeddings and gating reduces returns by 22\%. Lastly, including polarity embeddings also harms performance, indicating that spatio-temporal data primarily drives success.

\paragraph{Spatio-Temporal Embedding:}
Following the component analysis, we evaluate the impact of different coordinate projection techniques (\cref{tab:ablation_embedding}). We compare random Fourier features against a learnable look-up table (LUT), standard positional encoding \cite{vaswani2017attention}, and a simple linear projection. Random Fourier features yield the highest and most consistent returns.
\begin{table}[t]
    \begin{minipage}{.49\linewidth}
    \vbox to 6\baselineskip{
        \caption{Interquartile mean, 95\% confidence intervals, and relative change in IQM of the episodic return on Event-CarEnv Racing when removing or replacing components of the FLEET pipeline.
        }
        \label{tab:ablation_module}
        \vfill
      }
      
        \resizebox{\linewidth}{!}{
         \begin{tabular}{l|ccc} \toprule
            \multirow{2}{*}{Component} & \multicolumn{3}{c}{Return $\uparrow$}\\ 
             & IQM & CI95 & Rel. Change \\\midrule
            FLEET (Full Pipeline) & \textbf{46.96} &[43.28, 49.03] & - \\
            FLEET (Polarity Embedding) & 41.33 &[38.07, 45.40] & -11\% \\
            FLEET (No Temporal Embedding) & 36.44 &[33.43, 41.35] & -22\% \\
            FLEET (Transformer Backbone)  & 35.30 & [30.12, 41.37] &-24\% \\
            FLEET (No Cross-Attention)  & 33.28 & [28.70, 35.97] & -29\%\\
            \bottomrule
        \end{tabular}
        }
        
    \end{minipage}%
    \begin{minipage}{0.01\linewidth}
        \hfill
    \end{minipage}
    \begin{minipage}{.49\linewidth}
      \centering
        \vbox to 6\baselineskip{
        \caption{
        Comparison of spatio-temporal embedding techniques for FLEET. Results report the IQM and 95\% CI for episodic returns. Random Fourier features achieve the most robust performance.
        }
        \label{tab:ablation_embedding}
        \vfill
        }
        
        \resizebox{\linewidth}{!}{
       
        \begin{tabular}{l|ccc} \toprule
            \multirow{2}{*}{Spatiotemporal Embedding} & \multicolumn{3}{c}{Return $\uparrow$}\\ 
             & IQM & CI95 & Rel. Change \\\midrule
            Random Fourier Features  & \textbf{46.96} &[43.28, 49.03] & - \\
            Learnable LUT Encoding & 43.30 & [38.07, 47.90] & -7\%\\ 
            Positional Encoding & 39.95 & [36.19, 43.73] & -14\%\\   
            Linear Encoding & 20.64 & [18.89, 23.17] & -56\%\\ \bottomrule 
        \end{tabular}
        }
    \end{minipage} 
\end{table}

\section{Conclusion}
In this paper, we introduced FLEET, a novel, end-to-end trainable event camera feature extractor designed explicitly for reinforcement learning. By leveraging random Fourier features and a Perceiver-style cross-attention mechanism, our architecture processes variable-length event sequences and maps them into an expressive, fixed-size latent representation. FLEET bypasses computationally wasteful sparse grid aggregations and inherently decouples the complexity of the model's backbone from the sensor's spatial resolution.\\
We evaluated FLEET against state-of-the-art RL feature extractors across multiple dynamic continuous control tasks, including a novel high-throughput Event-CarEnv autonomous driving benchmark and an Event Inverted Double Pendulum environment. Our extensive empirical evaluations demonstrate that FLEET not only substantially outperforms all baseline methods but also exhibits remarkable robustness to high observation and control frequencies. A setting where traditional grid-based representations severely degrade.\\
Future work will explore efficient event sampling and filtering techniques to intelligently reduce the raw event sequence length processed by the feature extractor. Furthermore, because current real-world event camera datasets are tailored for supervised learning, a critical next step for practical adoption will be closing the sim-to-real gap and actively addressing the physical deployment of these policies.


\section*{Acknowledgements}
This work was supported by the MWK of Lower Saxony within Hybrint\linebreak (VWZN4219) and LCIS (VWZN4704), the Deutsche Forschungsgemeinschaft (DFG) under Germany’s Excellence Strategy within the Cluster of Excellence PhoenixD (EXC2122) and Quantum Frontiers (EXC2123), the European Union under grant agreement no. 101136006 – XTREME.

%
%
\bibliographystyle{splncs04}
\bibliography{main}

\clearpage
\setcounter{page}{1}
\renewcommand{\thesection}{\Alph{section}}
\title{Supplementary Material:\\ FLEET: Token-Based Feature Extraction for Event Camera-based Reinforcement Learning} 
\author{}\authorrunning{}\institute{}\email{}
\titlerunning{Supplementary Material: FLEET}
\maketitle

\setcounter{table}{5}
\setcounter{figure}{6}
\setcounter{equation}{5}

\section{Experiments}
\subsection{Experimental Setup}
All experiments were performed using Proximal Policy Optimization~\cite{schulman2017proximal} with a learning rate of $\lambda=0.0003$, employing separate MLP architectures for the actor and critic, each consisting of two hidden layers of 256 units.
The event data is processed by the respective feature extractor while the state vector of the environment is always processed by a single linear layer followed by a relu activation. The respective latent representations are concatenated prior to being passed to the agent. Following \cite{kostrikov2020image}, the feature extractors are shared between actor and critic.
Unless specified otherwise, the experiments were conducted without framestacking.
The agent was trained for $10^6$ steps with evaluation on 10 environment instances taking place every $10^5$ steps.\\
All methods were implemented in JAX~\cite{jax2018github} using the Flax~\cite{flax2020github} framework.
\subsection{Hyperparameters}
\cref{tab:parameters} and \cref{tab:baseline_parameters} provide the hyperparameters used in our experiments unless specified otherwise.
\begin{table}[htb]
\caption{FLEET hyperparameters used in all of our experiments.}
\label{tab:parameters}
    \begin{center}
        \begin{tabular}{l|c} \toprule
            Hyperparameter & Value\\\midrule
            \multicolumn{2}{l}{\textbf{FLEET}} \\ \midrule
            
            Event Sequence Length ($M$) & 2048  \\ 
            Embedding Dimension ($d_\text{embed}$) & 128 \\ 
            $\sigma_\text{temporal}$ & 5.0\\
            $\sigma_\text{spatial}$ & 20.0\\
            Number of Latent Tokens ($N$) & 32   \\ 
            Attention Heads ($h$) & 4\\
            Encoder MLP Widths & $(2 \cdot d_\text{embed}, 2 \cdot d_\text{embed}, d_\text{embed})$ \\
            \bottomrule 
        \end{tabular}
    \end{center}
\end{table}

\begin{table}[htb]
\caption{Hyperparameters for the baseline feature extractors used in our experiments.}
\label{tab:baseline_parameters}
    \begin{center}
        \begin{tabular}{l|c} \toprule
            Hyperparameter & Value\\\midrule
            
            \multicolumn{2}{l}{\textbf{NatureCNN}} \\ \midrule
            Conv. Architecture (Channels, Kernel, Stride) & $(32, 8, 1), (64, 4, 2), (64, 3, 1)$ \\
            Output Dimension & 512 \\ \midrule
            
            \multicolumn{2}{l}{\textbf{ImpalaCNN}} \\ \midrule
            Conv. Arch. (Features, Kernel, Stride, ResBlocks) & $(16, 3, 1, 2), (32, 3, 1, 2), (32, 3, 1, 2)$ \\
            Output Dimension & 256 \\ \midrule
            
            \multicolumn{2}{l}{\textbf{DMRCNN}} \\ \midrule
            Conv. Architecture (Channels, Kernel, Stride) & $(64, 5, 2), (128, 3, 2), (256, 3, 2), (256, 3, 2)$ \\
            Output Dimension & 3072 \\ \midrule
            
            \multicolumn{2}{l}{\textbf{LFF-DS}} \\ \midrule
            Item Encoder Units & $(128, 128)$ \\
            Set Encoder Units & $(256, 256, 128)$ \\ \midrule
            
            \multicolumn{2}{l}{\textbf{eVAE}} \\ \midrule
            Latent Dimension & 8 \\
            ECN Layers & $(64, 128, 1024)$ \\
            Encoder Layers & $(256, 16)$ \\
            Decoder Layers & $(128, 256)$ \\
            \bottomrule 
        \end{tabular}
    \end{center}
\end{table}

\subsection{Event-CarEnv}
\subsubsection{Event Generation}
We model the event camera dynamics of Event-CarEnv by converting rendered RGB frames first to grayscale frames $I$ and then into a logarithmic luminance space $L(\mathbf{u}, t) = \ln(I(\mathbf{u}, t) + \epsilon)$. Events are triggered asynchronously at pixel $\mathbf{u}$ whenever the log-intensity deviation $\Delta L$ from the stored reference state $L_{\text{ref}}$ exceeds the specific contrast threshold for that polarity. We define separate thresholds $\vartheta_{\text{on}}$ for increasing intensity ($\Delta L > 0$) and $\vartheta_{\text{off}}$ for decreasing intensity ($\Delta L < 0$). We adopt the calibrated parameter values from ESIM~\cite{rebecq2018esim}, where thresholds are sampled from normal distributions with means $\mu_{\text{on}}=0.55$, $\mu_{\text{off}}=0.41$ and standard deviation $\sigma=0.021$.\\
The polarity of the event is defined as $p = \text{sgn}(\Delta L)$. Let the effective threshold be $\vartheta = \vartheta_{\text{on}}$ if $p=1$ and $\vartheta = \vartheta_{\text{off}}$ if $p=-1$. Given a discrete simulation timestep $\Delta t$, the number of triggered events is $\mathcal{N} = \lfloor |\Delta L| / \vartheta \rfloor$. Following the interpolation approach of ESIM, we approximate the continuous-time nature of event generation within discrete rendering steps by estimating the timestamp $t_k$ for the $k$-th event:
\begin{equation}
    t_k = t_{\text{start}} + \left( \frac{k \cdot \vartheta \cdot p}{\Delta L} \right) \Delta t, \quad k \in \{1, \dots, \mathcal{N}\}\,.
\end{equation}
To ensure sensor realism, timestamps are constrained by a hardware refractory period $\tau_{\text{ref}}$, such that $t_k \leftarrow \max(t_k, t_{k-1} + \tau_{\text{ref}})$, and the reference state $L_{\text{ref}}$ is updated incrementally by $N \cdot \vartheta \cdot p$ to preserve sub-threshold residuals across simulation frames.\\

\subsubsection{Racing}
\begin{figure}[tb]
  \centering
  \includegraphics[width=0.9\textwidth]{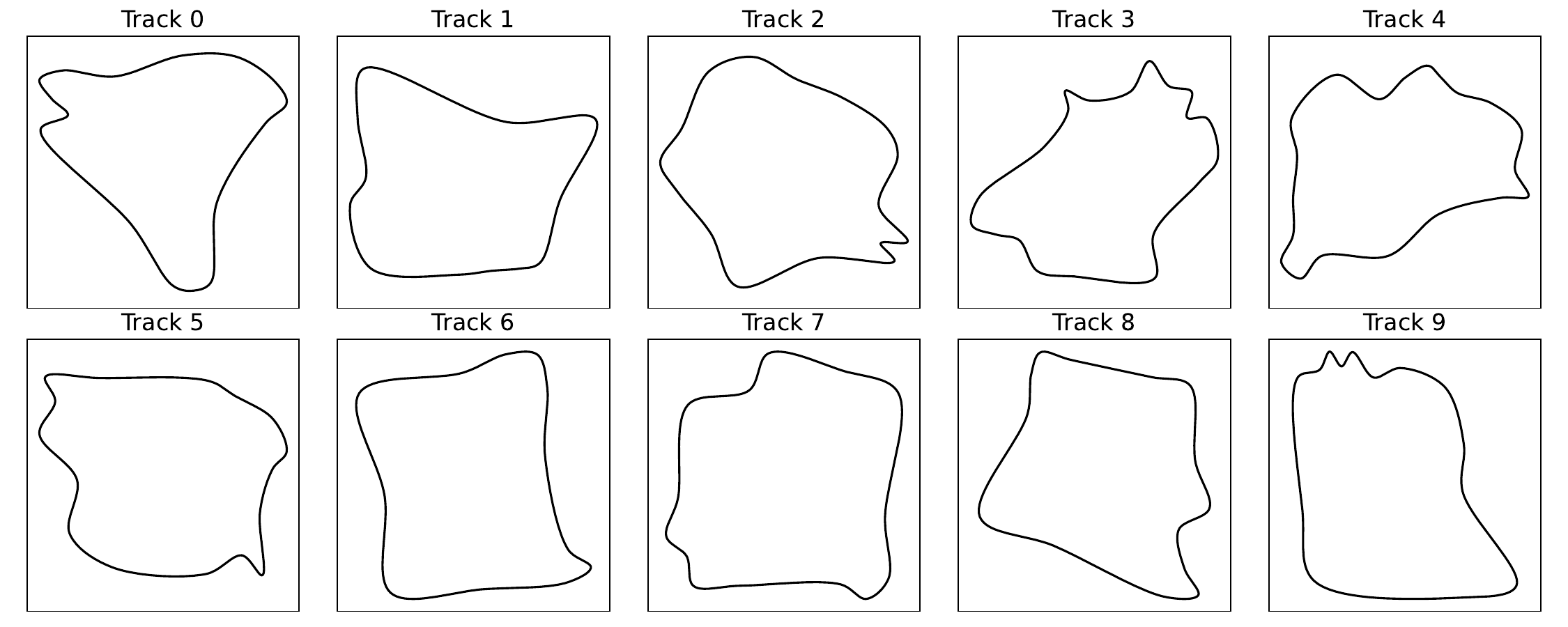}
  \caption{
  Course of the ten tracks used during evaluation in the Event-CarEnv Racing setting. The tracks cover different degrees of difficulty from easy (Track 6) to very hard (Track 3).
  }
  \label{fig:tracks}
\end{figure}

\begin{figure}[tb]
  \centering
  \begin{subfigure}{0.45\linewidth}
    \centering
    \includegraphics[width=1.0\textwidth]{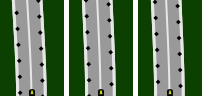}
    \caption{RGB Camera View}
    \label{fig:car_env_racing_straight_rgb}
  \end{subfigure}
  \hfill
  \begin{subfigure}{0.45\linewidth}
    \centering
    \includegraphics[width=1.0\textwidth]{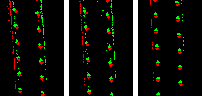}
    \caption{Event Camera View (accumulated)}
    \label{fig:car_env_racing_straight_event}
  \end{subfigure}
  \caption{Additional Sequence of RGB (\cref{fig:car_env_rgb}) and event (\cref{fig:car_env_event}) camera observations of a relatively straight section in the Event-CarEnv Racing scenario. Note that the event camera data has been temporarily aggregated into an image format strictly for human interpretability.}
  \label{fig:car_env_racing_straight}
\end{figure}

For a better understanding of the complexity of the task we provide the courses of the tracks used during evaluation the Racing setting in \cref{fig:tracks}.
The tracks differ in difficulty, with some requiring the agent to perform multiple sharp turns in short succession (\eg Tracks 0, 2 and 9) and others being easier due to a more simplistic track layout (\eg Track 6).\\
Furthermore, \cref{fig:car_env_racing_straight} visualizes a three frame sequence of the vehicle driving on a relatively straight segment of the track.\\
The reward function remains identical to \cite{schier2023learned}:
\begin{equation}
    r(\mathbf{s}, \mathbf{a}, \mathbf{s}') =-\mathds{1}_{\text {fail }}-0.2 \cdot \mathds{1}_{\text {collision }}+0.01 \cdot \mathbf{n}_{\text {track }} \cdot \mathbf{v}_{\text {ego }},
\end{equation}
with $\mathbf{v}_{\text {ego }}$ being the vehicles velocity vector which is projected onto the tracks forward direction $\mathbf{n}_{\text {track }}$.

\subsubsection{Parking}
\begin{figure}[tb]
  \centering
  \begin{subfigure}{0.45\linewidth}
    \centering
    \includegraphics[width=1.0\textwidth]{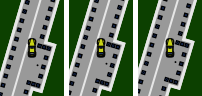}
    \caption{RGB Camera View}
    \label{fig:car_env_parking_rgb}
  \end{subfigure}
  \hfill
  \begin{subfigure}{0.45\linewidth}
    \centering
    \includegraphics[width=1.0\textwidth]{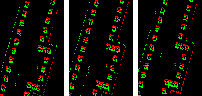}
    \caption{Event Camera View (accumulated)}
    \label{fig:car_env_parking_event}
  \end{subfigure}
  \caption{Sequence of RGB (\cref{fig:car_env_rgb}) and event (\cref{fig:car_env_event}) camera observations in the Event-CarEnv Parking scenario. Note that the event camera data has been temporarily aggregated into an image format strictly for human interpretability.}
  \label{fig:car_env_parking}
\end{figure}

\cref{fig:car_env_parking} shows a sequence of frames from the RGB camera and accumulated event camera data for the Event-CarEnv Parking scenario.\\
Similiar to the Racing scenario the reward function is kept unchanged from \cite{schier2023learned}:
\begin{equation} 
    r(\mathbf{s}, \mathbf{a}, \mathbf{s}')=-\mathds{1}_{\text {collision }}+c \cdot r_{\text {pose }}(\mathbf{s}, \mathbf{a}, \mathbf{s}'),
\end{equation}
with
\begin{align}
    & r_{\text {pose }}(\mathbf{s}, \mathbf{a}, \mathbf{s}')=\max \left(0,1-\frac{\vert x-u\vert}{x_{\max }}\right)\\ 
    & \cdot \max \left(0,1-\frac{\vert y-v\vert}{y_{\max }}\right) \cdot \max (0, \cos (\alpha-\kappa)). 
\end{align}
The second reward term $r_{\text {pose }}$ penalizes the current vehicle pose $(x, y, \alpha)$ deviating from the target pose $(u,v,\kappa)$. The scaling coefficients remain unchanged at $c=0.05$, $x_\text{max}=$ \si{40}{m} and $y_\text{max}=$ \si{5}{m}.

\subsection{Event Inverted Double Pendulum}
\begin{figure}[tb]
  \centering
  \begin{subfigure}{0.45\linewidth}
    \centering
    \includegraphics[width=1.0\textwidth]{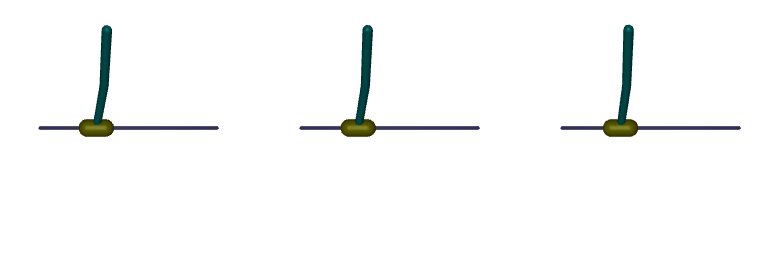}
    \caption{RGB Camera View}
    \label{fig:idp_rgb}
  \end{subfigure}
  \hfill
  \begin{subfigure}{0.45\linewidth}
    \centering
    \includegraphics[width=1.0\textwidth]{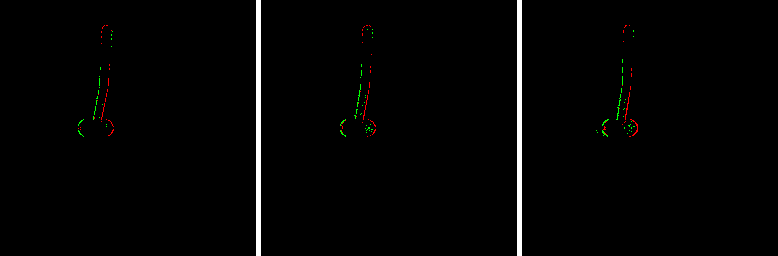}
    \caption{Event Camera View (accumulated)}
    \label{fig:idp_event}
  \end{subfigure}
  \caption{Sequence of RGB (\cref{fig:idp_rgb}) and event (\cref{fig:idp_event}) camera observations in the Inverted Double Pendulum. Note that the event camera data has been temporarily aggregated into an image format strictly for human interpretability.}
  \label{fig:idp}
\end{figure}

For the Inverted Double Pendulum environment the default frame-skipping mechanism is removed, resulting in a dense simulation step size of $\Delta t=0.01$ instead of $\Delta t=0.05$. 
To ensure comparability between the modified version of Inverted Double Pendulum used in the evaluation and the standard version, the discount factor, episode length and number of training steps is adjusted. The maximum episode length is proportionally increased to $5000$ interactions and the total number of environment steps to $5\cdot10^6$. The reward is augmented with a scaling factor $c_\text{scaling}$ based on $\Delta t$, keeping the episodic returns in the same range as the default setting. Thus the reward function can be formulated as:
\begin{equation}
    r(\mathbf{s}, \mathbf{a}, \mathbf{s}') = c_\text{scaling}\cdot(10\cdot\mathds{1}_\text{alive} - (0.01 \cdot x_\text{tip}^2 + (y_\text{tip}-2)^2) - (10^{-3}\cdot \omega_1^2 + 5\cdot10^{-3}\cdot\omega_2^2)),
\end{equation}
where $c_\text{scaling}=\Delta t/0.05$, $x_\text{tip}$ and $y_\text{tip}$ are the positions of the free swinging tip of the pendulum and $\omega_1$ and $\omega_2$ are the hinge's angular velocities.\\
Furthermore, we increase the discount factor to account for the increased episode length. In addition to the visual observation of the environment, the observation space is augmented with the velocity of the cart and angular velocities of the pendulum~\cite{towers2024gymnasium}.\\
In contrast to the Event-CarEnv scenarios, the Inverted Double Pendulum setting keeps the cameras position static as shown in \cref{fig:idp}. While in Racing and Parking the sensors only capture a part of the track, in Inverted Double Pendulum the complete environment is visible at all times.

\section{Memory Usage and Performance} \label{sec:performance_memory}
\begin{figure}[tb]
  \centering
  \begin{subfigure}{0.45\linewidth}
    \centering
    \includegraphics[width=1.0\textwidth]{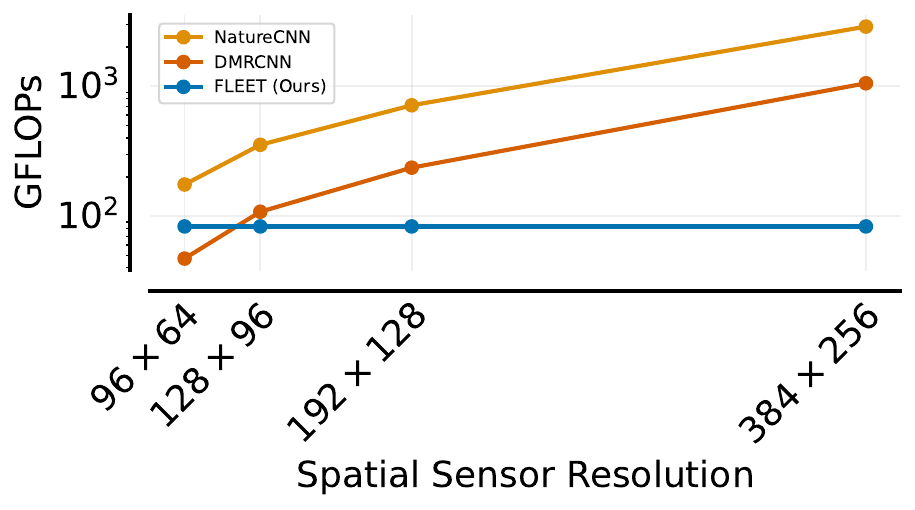}
    \vbox to 2\baselineskip{\caption{Computational load per minibatch in GFLOPs.}
    \label{fig:performance}}
    
  \end{subfigure}
  \hfill
  \begin{subfigure}{0.45\linewidth}
    \centering
    \includegraphics[width=1.0\textwidth]{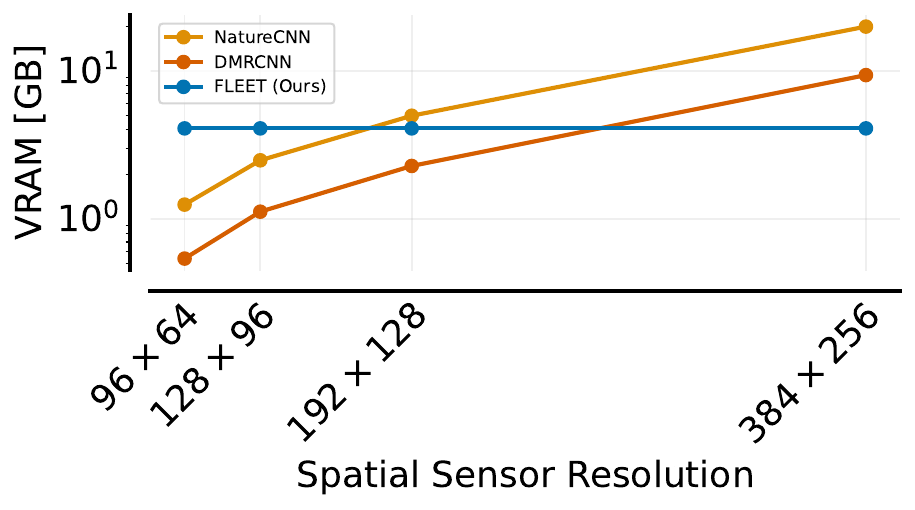}
    \vbox to 2\baselineskip{\caption{Memory required per minibatch in GB.}
    \label{fig:memory}}
    
  \end{subfigure}
  \caption{
  Comparison of computational cost (\cref{fig:performance}) and memory consumption (\cref{fig:memory}) per minibatch (5120 transitions) for NatureCNN, DMRCNN, and FLEET across increasing spatial resolutions. 
  }
  \label{fig:performance_vram}
\end{figure}

A core advantage of the FLEET architecture over standard CNN-based approaches is its ability to maintain superior performance as spatial resolution scales up. While baseline methods like NatureCNN can also yield increased episodic returns at higher spatial resolutions, they do so at a severe computational penalty.
To quantify this bottleneck, the computational load in floating point operations (FLOPs), the execution time, and the VRAM usage per minibatch (5120 transitions) were analyzed across four spatial sensor resolutions ($96\times64$, $128\times96$, $192\times128$, and $384\times256$) in \cref{tab:memory_performance}, \cref{tab:execution_time}, and \cref{fig:performance_vram}.
All experiments were conducted on a dedicated compute node equipped with an NVIDIA RTX 3090 GPU, 4 CPU threads, and 32 GB of system RAM.
At the lowest resolution of $96\times64$, FLEET requires $83.28$ GFLOPs, which is initially higher than DMRCNN at $47.11$ GFLOPs. However, despite this higher theoretical FLOP count, FLEET exhibits a significantly lower inference execution time of \SI{11.76}{ms} compared to DMRCNN's \SI{81.48}{ms} and NatureCNN's \SI{286.63}{ms}. This discrepancy demonstrates that FLEET relies on highly parallelizable, GPU-friendly operations, such as dense matrix multiplications in its attention mechanisms, which maps efficiently to modern hardware..
After only a slight increase in spatial resolution to $128\times96$, FLEET immediately demands the lowest computational load among all tested models. At this $128\times96$ resolution, FLEET's constant $83.28$ GFLOPs drops below DMRCNN's $108.05$ GFLOPs and requires roughly 23\% of the compute needed by NatureCNN's $353.93$ GFLOPs. NatureCNN's high computational load can be explained by the choice to decrease the stride to $1$ instead of $4$ to prevent the feature extractor from skipping single-pixel events.
As resolutions scale further, this efficiency gap widens exponentially because CNN architectures must process dense spatial grids, causing their compute and memory requirements to scale quadratically. FLEET completely bypasses this representation degradation and computational bottleneck by compressing the variable event stream into a fixed number of latent tokens via cross-attention. Consequently, FLEET maintains a strictly constant load of $83.28$ GFLOPs and a fixed \SI{4.09}{GB} VRAM footprint across all evaluated resolutions.
At the maximum evaluated resolution of $384\times256$, FLEET uses just 8\% of the $1053.09$ GFLOPs required by DMRCNN, and a mere 3\% of the $2876.33$ GFLOPs demanded by NatureCNN. In terms of actual inference speed, this translates to FLEET executing over 310 times faster than DMRCNN (\SI{3690.64}{ms}) and over 400 times faster than NatureCNN (\SI{4778.59}{ms}) per minibatch at this resolution. This completely flat scaling curve should enable our architecture to efficiently leverage modern, high-definition event sensors without the immense overhead associated with grid-based methods.
\begin{table}[tb]
\caption{
Comparison of computational cost (GFLOPs) and memory consumption (VRAM in GB) per minibatch (5120 transitions) for NatureCNN, DMRCNN, and FLEET across increasing spatial resolutions. Because FLEET directly processes a fixed-length sequence of events rather than sparse grids, its computational load and memory footprint remain perfectly constant, effectively decoupling inference cost from the sensor's spatial resolution.}
\label{tab:memory_performance}
    \begin{center}
    \resizebox{\columnwidth}{!}{%
        \begin{tabular}{l|cc|cc|cc|cc} \toprule
            \multirow{2}{*}{Feature Extractor} & \multicolumn{2}{c|}{$96\times 64$}  & \multicolumn{2}{c|}{$128\times 96$}  & \multicolumn{2}{c|}{$192\times 128$} & \multicolumn{2}{c}{$384\times 256$} \\
             & GFLOPs($\downarrow$) &  VRAM [GB]($\downarrow$) & GFLOPs($\downarrow$) &  VRAM [GB]($\downarrow$)  & GFLOPs($\downarrow$) &  VRAM [GB]($\downarrow$)  & GFLOPs($\downarrow$) &  VRAM [GB]($\downarrow$)  \\\midrule
             NatureCNN &  175.11 & \underline{1.25} & 353.93 & \underline{2.49} & 712.84 & 4.98 & 2876.33 & 19.92\\
             DMRCNN &  \textbf{47.11} & \textbf{0.54} & \underline{108.05} & \textbf{1.12} & \underline{236.07} & \textbf{2.28} & \underline{1053.09} & \underline{9.37}\\
             FLEET (\textbf{ours}) &  \underline{83.28} & 4.09 & \textbf{83.28} & 4.09 & \textbf{83.28} & \underline{4.09} & \textbf{83.28} & \textbf{4.09}\\\bottomrule 
        \end{tabular}
    }
    \end{center}
\end{table}

\begin{table}[tb]
    \centering
    \caption{Inference execution time per minibatch for NatureCNN, DMRCNN and FLEET across increasing spatial resolutions. FLEET leverages GPU-friendly operations, leading to lower execution times than it's competitors.}
    \label{tab:execution_time}
    \resizebox{\columnwidth}{!}{%
        \begin{tabular}{l|c|c|c|c} \toprule
            \multirow{2}{*}{Feature Extractor} & $96\times 64$  & $128\times 96$  & $192\times 128$ & $384\times 256$ \\
             & Exec. Time [\si{ms}]($\downarrow$) & Exec. Time [\si{ms}]($\downarrow$)  & Exec. Time [\si{ms}]($\downarrow$)  & Exec. Time [\si{ms}]($\downarrow$) \\\midrule
             NatureCNN &  286.63 & 577.67 & 1165.83 & 4778.59\\
             DMRCNN &  \underline{81.48} & \underline{260.72} & \underline{644.07} & \underline{3690.64}\\
             FLEET (\textbf{ours}) & \textbf{11.76} & \textbf{11.76} & \textbf{11.76} & \textbf{11.76}\\\bottomrule 
        \end{tabular}
    }
\end{table}

\section{Ablations}
This section includes ablations of our feature extractor that could not be included in the main submission due to page constraints.

\paragraph{Sequence Sampling Technique:}
We evaluate the impact of our sequence sampling strategy by comparing it against a temporal sampling baseline that simply selects the most recent $M$ events. As shown in Table \ref{tab:ablation_sequence_sampling}, replacing FLEET's stratified sampling with this temporal heuristic results in a 21\% drop in episodic return. While dense scenes inevitably require discarding events to satisfy sequence length constraints, our approach distributes this reduction uniformly across the entire sampling period. This strategy successfully preserves the overall spatio-temporal geometry of the scene and avoids the severe temporal blindspots that occur when truncating earlier events. Maintaining this structural integrity provides a critical advantage for the agent, as real event camera bursts are highly asynchronous and rarely linearly spaced in time. 
\begin{table}
\caption{Interquartile mean, 95\% confidence intervals, and relative change in IQM of the episodic return when using temporal sampling (choose last $M$ events) when compared to the sampling approach used in FLEET.}
\label{tab:ablation_sequence_sampling}
    \begin{center}
     \begin{tabular}{l|ccc} \toprule
            \multirow{2}{*}{Sequence Sampling Technique} & \multicolumn{3}{c}{Return $\uparrow$}\\
             & IQM & CI95 & Rel. Change \\\midrule
            FLEET (\textbf{ours}) & \textbf{46.96} &[43.28, 49.03] & - \\
            Temporal Sampling & 37.02 &[34.68, 39.89] & -21\% \\
            \bottomrule
        \end{tabular}
    \end{center}
\end{table}

\paragraph{Matched Sequence Baseline:}
To isolate the performance gains of our proposed architecture from the event sequence sampling, we evaluated a parameter-matched MLP sequence baseline. This baseline processes identical sampled event sequences but replaces the Random Fourier Features with standard linear embeddings and entirely removes the cross-attention bottleneck. As shown in Table \ref{tab:ablation_matched_sequence}, eplacing FLEET with this baseline results in a 35\% drop in episodic return. While an alternative Transformer-based sequence encoder could theoretically be applied, its computational complexity scales quadratically with the number of events in the absence of our bottleneck mechanism, which undermines the efficiency required for high-frequency control. These results demonstrate that while our sampling strategy provides a strong foundation, FLEET's specific architectural innovations account for approximately one-third of the overall performance improvement.

\begin{table}
\caption{Interquartile mean, 95\% confidence intervals, and relative change in IQM of the episodic return comparing the full FLEET architecture to a parameter-matched MLP sequence baseline. This comparison isolates the performance gains provided by the Random Fourier Features and the cross-attention bottleneck from the inherent benefits of the underlying sequence sampling strategy.}
\label{tab:ablation_matched_sequence}
    \begin{center}
     \begin{tabular}{l|ccc} \toprule
            \multirow{2}{*}{Sequence Sampling Technique} & \multicolumn{3}{c}{Return $\uparrow$}\\
             & IQM & CI95 & Rel. Change \\\midrule
            FLEET (\textbf{ours}) & \textbf{46.96} &[43.28, 49.03] & - \\
            Matched Sequence Baseline & 30.06 &[26.13, 36.12] & -35\% \\
            \bottomrule
        \end{tabular}
    \end{center}
\end{table}

\begin{figure}[tb]
  \centering
  \begin{subfigure}{0.45\linewidth}
    \centering
    \includegraphics[width=1.0\textwidth]{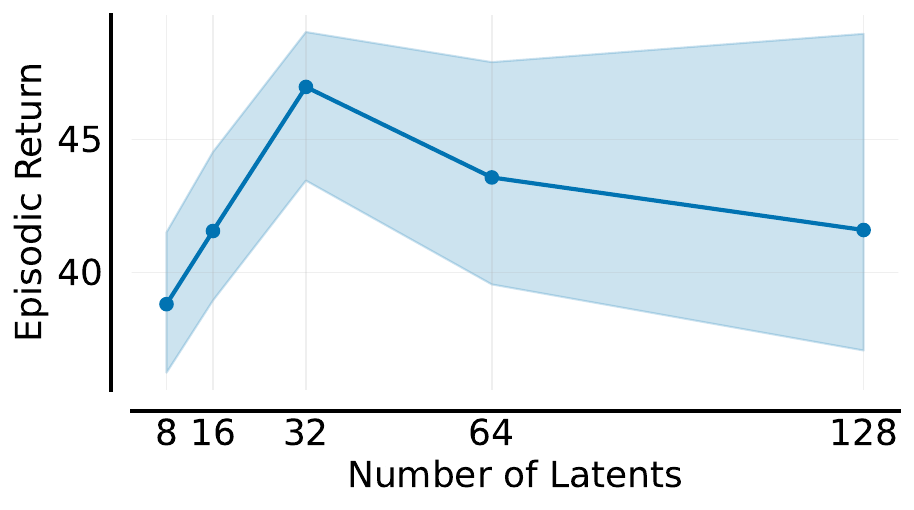}
    \vbox to 6\baselineskip{\caption{Results of FLEET utilizing different numbers of latent tokens $N \in \{8, 16, 32, 64, 128\}$. We report the interquartile mean and 95\% confidence intervals of the episodic return. The data indicates that $N=32$ latent tokens strikes the optimal balance necessary to sufficiently capture the dynamics of the scenarios.}
    \label{fig:ablations_latents}}
    
  \end{subfigure}
  \hfill
  \begin{subfigure}{0.45\linewidth}
    \centering
    \includegraphics[width=1.0\textwidth]{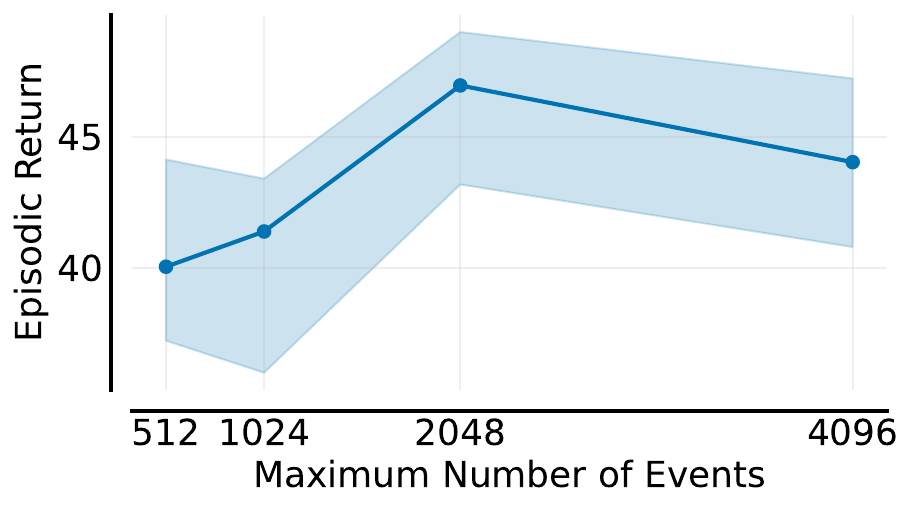}
    \vbox to 6\baselineskip{\caption{IQM and CI95 of the episodic return for varying number of events $M\in \{512, 1024, 2048, 4096\}$ after the event sequence sampling module. Selecting a too small $M$ can lead to a reduction in performance.}
    \label{fig:ablation_seqlen}}
    
  \end{subfigure}
  \caption{
  Hyperparameter study on the Event-CarEnv Racing scenario for the number of latents $N$ in the cross-attention module (\cref{fig:ablations_latents}) and the maximum event sequence length $M$ output by the event sequence sampling module (\cref{fig:ablation_seqlen}). 
  }
  \label{fig:ablatin_latents_seqlen}
\end{figure}

\paragraph{Number of Latent Tokens $N$:}
As demonstrated before, the cross-attention mechanism successfully maps raw events to semantically consistent latent tokens.  Therefore, we investigate how varying the total number of these latent tokens $N$ impacts the episodic return. We evaluate FLEET across $N \in \{8, 16, 32, 64, 128\}$, with the results visualized in \cref{fig:ablations_latents}.
The experiments reveal that decreasing the number of latent tokens below $N=32$ leads to a steady decline in episodic return. Conversely, increasing the token count beyond $N=32$ not only lowers the average return but also widens the confidence intervals.\\
We hypothesize that with too few latent tokens, the feature extractor loses expressivity, forcing the model to compress semantically distinct events into the same token. On the other hand, an excess of tokens causes the model to distribute a single semantic event across multiple latents. This introduces ambiguity during the learning process, which manifests as higher variance in the confidence intervals.
\begin{table}[tb]
\caption{IQM and CI95 of FLEET on Event-CarEnv Racing with varying number of latents $N$ in the cross-attention module. Both too many and not enough latents lead to performance degradation.}
\label{tab:ablation_num_latents}
    \begin{center}
    \resizebox{\columnwidth}{!}{%
        \begin{tabular}{l|r @{\,\,} r @{\,} r |r @{\,\,} r @{\,} r |r @{\,\,} r @{\,} r |r @{\,\,} r @{\,} r |r @{\,\,} r @{\,} r} \toprule
            \multirow{3}{*}{Feature Extractor} & \multicolumn{3}{c|}{$N=8$}  & \multicolumn{3}{c|}{$N=16$}  & \multicolumn{3}{c|}{$N=32$} & \multicolumn{3}{c|}{$N=64$} & \multicolumn{3}{c}{$N=128$}\\
            & \multicolumn{3}{c|}{(Return $\uparrow$)} & \multicolumn{3}{c|}{(Return $\uparrow$)} & \multicolumn{3}{c|}{(Return $\uparrow$)} & \multicolumn{3}{c|}{(Return $\uparrow$)} & \multicolumn{3}{c}{(Return $\uparrow$)}\\
             & IQM &  \multicolumn{2}{c|}{CI95} & IQM &  \multicolumn{2}{c|}{CI95} & IQM &  \multicolumn{2}{c|}{CI95} & IQM &  \multicolumn{2}{c|}{CI95} & IQM &  \multicolumn{2}{c}{CI95}\\\midrule
            FLEET (\textbf{ours})& 38.81 &[35.99, & 41.09] & 41.55 &[38.94, & 44.57]& \textbf{46.96} &[43.28,& 49.03]& 43.56 & [39.51, & 47.71] & 41.59 & [37.11, & 48.52]\\\bottomrule 
        \end{tabular}
    }
    \end{center}
\end{table}

\paragraph{Event Sequence Length $M$:}
Since the number of events $D$ within a sampling period is variable due to the asynchronous nature of the sensor, the raw event sequence can be difficult to process. CNN-based approaches avoid this by aggregating the data into a grid which can result in a loss of information. FLEET works directly on the event sequence data. However, the size of the attention matrix of the cross-attention module is dependent on the number of events. Furthermore, choosing JAX for our implementation requires fixed input sizes to leverage JAX's acceleration. Thus FLEET preprocesses the variable length raw event sequence to a fixed-length sequence. For this, the number of events is padded with zero values if there are fewer than $M$ events and sampled if there are more than $M$ events.\\
In this experiment, we investigate the impact of different sequence lengths $M$ in the Event-CarEnv Racing scenario. The scenario has a mean raw event count of $3741$ events per observation step.
\cref{tab:ablation_seqlen} and \cref{fig:ablation_seqlen} show that reducing $M$ too much can lead to a drop in episodic return as too few relevant events are captured.
\begin{table}[tb]
\caption{Interquartile mean (IQM) and 95\% confidence intervals (CI95) of the episodic return for the FLEET architecture on the Event-CarEnv Racing scenario, evaluated across varying event sequence lengths $M$. The results illustrate the impact of the sequence length sampled prior to the cross-attention bottleneck, demonstrating how excessively small values of $M$ can constrain the network's performance, while still performing better than CNN-based approaches.}
\label{tab:ablation_seqlen}
    \begin{center}
    \resizebox{\columnwidth}{!}{%
        \begin{tabular}{l|r @{\,\,} r @{\,} r |r @{\,\,} r @{\,} r |r @{\,\,} r @{\,} r |r @{\,\,} r @{\,} r } \toprule
            \multirow{3}{*}{Feature Extractor} & \multicolumn{3}{c|}{$M=512$}  & \multicolumn{3}{c|}{$M=1024$}  & \multicolumn{3}{c|}{$M=2048$}  & \multicolumn{3}{c}{$M=4096$}\\
            & \multicolumn{3}{c|}{(Return $\uparrow$)} & \multicolumn{3}{c|}{(Return $\uparrow$)} & \multicolumn{3}{c|}{(Return $\uparrow$)} & \multicolumn{3}{c}{(Return $\uparrow$)}\\
             & IQM &  \multicolumn{2}{c|}{CI95} & IQM &  \multicolumn{2}{c|}{CI95} & IQM &  \multicolumn{2}{c|}{CI95} & IQM &  \multicolumn{2}{c}{CI95}\\\midrule
            FLEET (\textbf{ours})& 40.05 &[37.23, & 44.14] & 41.39 &[36.00, & 43.41]& \textbf{46.96} &[43.28,& 49.03]& 44.04 & [40.80, & 47.23]\\\bottomrule 
        \end{tabular}
    }
    \end{center}
\end{table}

\paragraph{Embedding Aggregation Technique:}
FLEET performs element-wise gating of the spatial embedding with a linear projection of the temporal embedding. In this ablation we replace the temporal gating with other aggregation techniques. We perform element-wise addition and multiplication. Furthermore, we also try concatenating the different embeddings and projecting them back to the original embedding dimension afterwards. The results can be found in \cref{tab:ablation_aggregation}. Using the temporal gating mechanism instead of the other approaches results in higher episodic returns, strengthening the motivation to use temporal gating in our approach.
\begin{table}
\caption{Interquartile mean, 95\% confidence intervals, and relative change in IQM of the episodic return when replacing the temporal gating embedding aggregation strategy with element-wise addition or multiplication and concatenation. Temporal gating shows the highest episodic returns.}
\label{tab:ablation_aggregation}
    \begin{center}
     \begin{tabular}{l|ccc} \toprule
            \multirow{2}{*}{Aggregation Technique} & \multicolumn{3}{c}{Return $\uparrow$}\\ 
             & IQM & CI95 & Rel. Change \\\midrule
            Temporal Gating & \textbf{46.96} &[43.28, 49.03] & - \\
            Addition & 40.03 &[36.47, 44.25] & -14\% \\
            Multiplication & 31.83 &[27.39, 35.80] & -32\% \\
            Concatenation  & 39.74 & [34.35, 45.91] & -15\%\\
            \bottomrule
        \end{tabular}
    \end{center}
\end{table}

\end{document}